\documentclass[10pt,twocolumn,letterpaper]{article}

\usepackage[pagenumbers]{iccv} 

\usepackage{tabu}
\usepackage{booktabs}
\usepackage{times}
\usepackage{epsfig}
\usepackage{graphicx}
\usepackage{amsmath}
\usepackage{amssymb}
\usepackage{algorithm}
\usepackage{color}
\usepackage{algorithmic}
\usepackage{bbm}
\usepackage{bm}
\usepackage{multirow}
\usepackage{comment}
\usepackage{here}
\usepackage{arydshln}
\usepackage{multicol}
\usepackage{iccv}
\usepackage{textcomp}
\usepackage{tipa}
\usepackage{url}
\usepackage{float}
\usepackage{appendix} 
\usepackage{tabularx}
\usepackage{pgffor}

\definecolor{iccvblue}{rgb}{0.21,0.49,0.74}

\usepackage[pagebackref,breaklinks,colorlinks,allcolors=iccvblue]{hyperref}

\title{Dataset Distillation via Vision-Language Category Prototype}

\author{Yawen Zou$^{1}$, Guang Li$^{2}$, Duo Su$^{3}$, Zi Wang$^{4}$, Jun Yu$^{4}$, Chao Zhang$^{1}$ \\
$^{1}$University of Toyama, 
$^{2}$Hokkaido University, 
$^{3}$Tsinghua University, 
$^{4}$Niigata University
}

\begin{document}
\maketitle

\begin{abstract}
Dataset distillation (DD) condenses large datasets into compact yet informative substitutes, preserving performance comparable to the original dataset while reducing storage, transmission costs, and computational consumption. However, previous DD methods mainly focus on distilling information from images, often overlooking the semantic information inherent in the data. The disregard for context hinders the model's generalization ability, particularly in tasks involving complex datasets, which may result in illogical outputs or the omission of critical objects. In this study, we integrate vision-language methods into DD by introducing text prototypes to distill language information and collaboratively synthesize data with image prototypes, thereby enhancing dataset distillation performance. 
Notably, the text prototypes utilized in this study are derived from descriptive text information generated by an open-source large language model. This framework demonstrates broad applicability across datasets without pre-existing text descriptions, expanding the potential of dataset distillation beyond traditional image-based approaches. Compared to other methods, the proposed approach generates logically coherent images containing target objects, achieving state-of-the-art validation performance and demonstrating robust generalization. Source code and generated data are available in \url{https://github.com/zou-yawen/Dataset-Distillation-via-Vision-Language-Category-Prototype/}.
\end{abstract}    
\section{Introduction}
The rapid development of deep learning, driven by powerful computational resources and extensive datasets, has posed substantial challenges for researchers due to the increasing demands for computational power and storage capacity \cite{lecun2015deep,deng2009imagenet,liu2025evolution}. Confronted with these issues, dataset distillation (DD) has emerged as a promising approach to extracting and distilling information from large datasets. DD synthesizes smaller datasets with high information density that approximate the performance of downstream tasks conducted on the original dataset \cite{wang2018datasetdistillation,kim2022dataset,sachdeva2023survey,geng2023survey}. Moreover, the surrogate dataset helps alleviate concerns about privacy and copyright issues \cite{dong2022privacy,carlini2022no,lei2023survey}. 

\begin{figure}[t] 
\centering 
\includegraphics[width=0.48\textwidth]{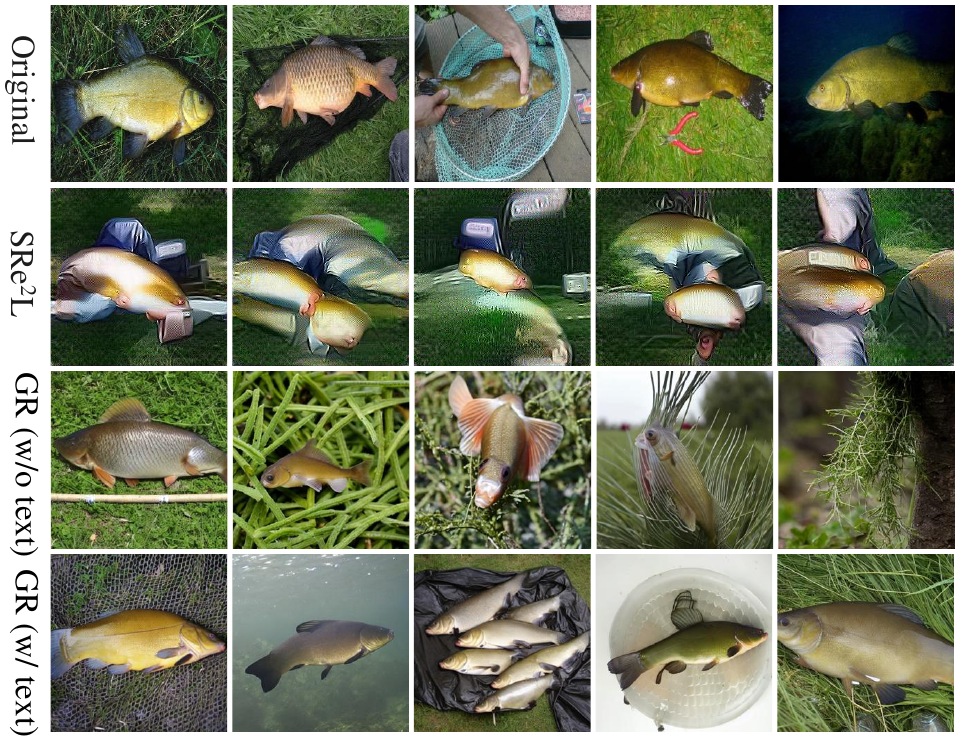} 
\caption{Visualization results of $\text{SRe}^2\text{L}$, GR (w/o text), and GR (w/ text). GR (w/o or w/ text) denotes the generative model outputs without and with text descriptions. Notably, GR (w/ text) captures rich details of target objects while preserving background diversity, leading to more comprehensive and visually coherent images.
} 
\label{fig1} 
\end{figure}

Dataset distillation was first introduced by Wang et al. \cite{wang2018datasetdistillation} and the follow-up studies have made significant advancements in recent years \cite{li2022awesome,zhou2022dataset,zhao2021datasetcondensation,zhao2023dataset,su2024d,gu2024efficient}. The earliest methods primarily include meta-learning based and matching based methods. In meta-learning methods, the distilled data are optimized as hyperparameters within a bi-level optimization framework \cite{deng2022remember,nguyen2021kip,loo2023dataset,sucholutsky2021soft,loo2022efficient}. And the matching based methods distill images via parameter matching \cite{kim2022dataset,lee2022dataset,liu2023dream,zhao2021dataset,cazenavette2022dataset} and distribution matching \cite{zhao2023dataset,wang2022cafe,zhao2022synthesizing,liu2022graph,zhao2023idm}.

However, these methods are compute-intensive and require substantial runtime due to iterative optimization, which ensures their representativeness. Recently, several studies utilize generative models \cite{wang2023dim,moser2024latent,gu2024efficient,su2024d} for DD to optimize latent features rather than image pixels, achieving faster training and improved performance. Gu et al. \cite{gu2024efficient} propose extra minimax criteria for diffusion models to generate representative and diverse synthetic data.
Su et al. \cite{su2024d} integrate the diffusion model into DD to extract embedding features, employing clustering centers as class prototypes, which are subsequently combined with label texts to generate images. Compared to traditional methods, generative models exhibit consistent GPU consumption across various images per class (IPC) settings while significantly reducing computational costs and demonstrating superior performance.

Despite the considerable progress in applying diffusion models to DD, these methods still face significant challenges. As illustrated in Fig. \ref{fig1}, existing methods (GR (w/o text)) occasionally generate images that consist only of background features without the target objects. Additionally, they struggle to generate logically coherent images, often producing unrealistic outputs such as dogs with five legs. Another critical issue is the co-occurrence bias inherent in the dataset, where certain objects or features frequently appear together. For example, fish and green plants often co-occur in surrogate data. This bias causes models to overemphasize these co-occurring features, prioritizing their coexistence over the accurate representation of individual elements. These challenges arise because existing methods focus solely on distilling information from image features while neglecting semantic information. As a result, they lack the necessary contextual understanding to generate coherent images, ultimately compromising the quality of the synthesized data.

In this work, we propose a novel framework that integrates vision-language methods into DD to improve the performance of the distilled datasets. Unlike traditional approaches that rely solely on visual features, our method leverages paired image-text representations to guide the generative process, enabling the generation of logically coherent and semantically enriched datasets. We first obtain image prototypes by applying K-means clustering to the features compressed from a pre-trained autoencoder, thereby capturing representative visual characteristics. Building on this foundation, we construct text prototypes for each cluster from textual descriptions generated by open-source large language models (LLMs). To ensure representativeness and diversity, words common across all clusters are excluded, as they fail to characterize individual clusters effectively. The sentence with the highest matching score to these feature words is selected as the final text prototype, ensuring an accurate representation of both the central theme and the unique characteristics of each cluster.

Compared with previous methods, our approach integrates both text and image prototypes to improve the performance of DD. Our method alleviates the previous issues mentioned in \cite{su2024d,gu2024efficient}, and the generated images contain the intended objects and exhibit logical coherence. Furthermore, the experimental results demonstrate that the distilled dataset outperforms the state-of-the-art methods in top-1 accuracy. Specifically, we observe improvements of 3.9\%, 4.9\%, and 3.5\% on ImageNette \cite{howardsmaller}, and 2.9\%, 4.2\%, and 2.5\% on ImageIDC \cite{kim2022dataset}, achieving superior performance over previous methods under IPC settings of 10, 20, and 50, respectively.
The source
code is provided in the supplementary material. 

The contributions of this study are summarized as follows:
\begin{itemize}
    \item To the best of our knowledge, this is the first work that integrates language information into visual dataset distillation for classification tasks. By leveraging textual descriptions, our approach enriches image-based information with crucial details such as shape, color, background, \etc, thereby mitigating existing limitations.
    \item We employ open-source large language models in DD to generate descriptive text for unimodal data, addressing the lack of textual descriptions in existing DD benchmarks and improving the generalization of our method.
    \item We propose a novel text prototype scheme for DD, which leverages word frequency within each cluster to ensure the representativeness and diversity of the text prototypes.
\end{itemize}

\begin{figure*}[htbp] 
    \centering
    \includegraphics[width=0.84\textwidth]{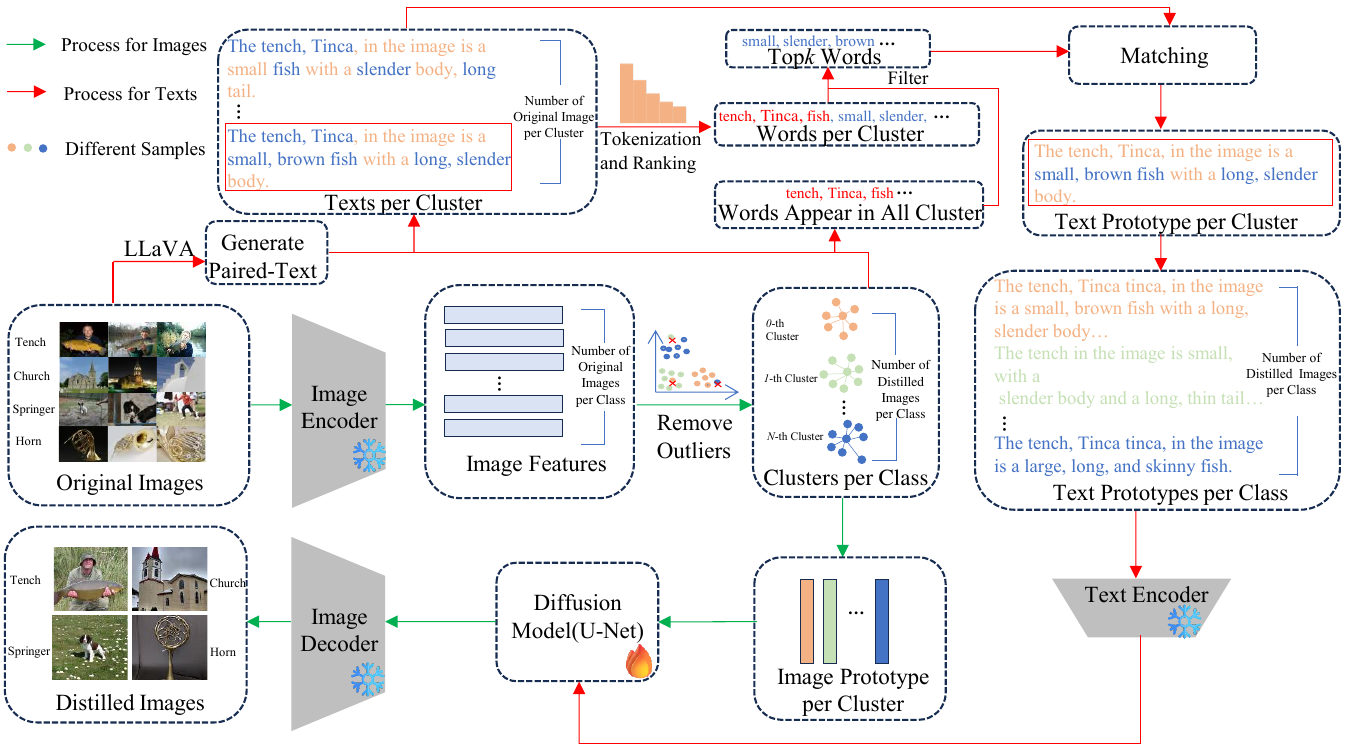} 
    \caption{Overview of the proposed framework. The framework starts with generating image-text pairs using the LLaVA model, followed by training a diffusion model. Image features are then compressed with an autoencoder, outliers are removed, and K-means clustering is applied to create image prototypes. For text prototypes, frequent words are extracted from descriptions, and the most representative sentence is selected. Finally, these prototypes guide the diffusion model to synthesize diverse and representative images.}
    \label{fig２}
\end{figure*}

\section{Preliminaries}
Dataset distillation aims to construct a small compact dataset $S = \left \{ \left ( X_{i},y_{i}  \right )  \right \} _{i=1}^{N_{S} }$ that encapsulates key information from a large-scale one $T = \left \{ \left ( X_{i},y_{i}  \right )  \right \} _{i=1}^{N_{T} }$, where $X_{i}$ represents an image, $y_{i}$ denotes its corresponding class label, and $N_{S} << N_{T}$ \cite{wang2018datasetdistillation,yu2023review}.
 By synthesizing such a dataset, DD enables models trained on 
$S$ to achieve comparable performance to those trained on 
$T$, while significantly reducing storage and computational costs.

Recently, diffusion models have emerged as powerful tools for generating high-quality synthetic data, making them a promising approach for dataset distillation. These models, known for their outstanding performance in generative tasks, synthesize high-quality images by adding Gaussian noise and then reversing the process to reconstruct them, ensuring consistency between input and output spaces. In this work, we employ Stable Diffusion \cite{rombach2022high,von-platen-etal-2022-diffusers} for training, which comprises three key components: an image encoder (VAE), a text encoder (CLIP), and a U-Net. The VAE consists of an encoder $E$ and a decoder $D$, where $E$ projects the image into a latent space $z = E(x)$, and $D$ reconstructs the latent code back to the image space $\hat{x} = D(z)$. The text encoder projects input prompts into the same feature space as the images, facilitating text-guided image generation. The U-Net utilizes a conditional model to process noisy latent $z_t$ and predicts the noise, incorporating both the timestep ($t$) and the text embedding for guidance. The training objective of diffusion models is defined as:

\begin{equation}
\mathcal{L_{DM}} = \lVert \epsilon_\theta(z_t, c) - \epsilon \rVert_2^2, 
\end{equation}

where $c$ is conditioning vector encoded with corresponding text, $\epsilon _{\theta } \left ( z_{t},c  \right ) $ is the predicted noise, and $\epsilon$ is the ground truth.

\section{Method}
In this section, we present a vision-language dataset distillation method designed to enhance the validation performance of synthetic datasets, as illustrated in Fig. \ref{fig２}. By incorporating visual and textual information, our method addresses the limitations of conventional approaches that rely solely on image data. The complete distillation process is outlined in Algorithm \ref{alg:1}.

\subsection{Paired Description Generation}\label{A}

In contrast to \cite{su2024d,gu2024efficient}, which distills information solely from images, we also distill information from description texts to enhance dataset quality. However, the existing benchmark datasets lack corresponding descriptive textual information, and annotating these data manually is time-consuming and labor-intensive. Fortunately, the rapid development of large language models has made this task feasible. Hence, we leverage the open-source large language model LLaVA \cite{liu2024llavanext,liu2024improved,liu2023visual} to generate the corresponding text by designing structured prompts. These descriptions capture additional semantic attributes, such as logical relationships and contextual details, which are not directly inferred from image features. The prompt is designed as follows:

Prompt = ``Describe the physical appearance of the \{\$CLASSNAME\} in the image. Include details about its shape, posture, color, and any distinct features."


\begin{algorithm}[h!]
\begin{algorithmic}[1]
\STATE \textbf{Input:} $(R, L)$: Real images and their labels, $LLM$: Large language model, $DM$: Diffusion model.
\STATE \textbf{Generate descriptive text}: $T = LLM(R)$
\STATE \textbf{DM training}: Fine-tune the DM using image-text pairs $(R,T)$. \bm{$E$}: Encoder, \bm{$D$}: Decoder, \bm{$\tau_\theta$}: Text encoder, \bm{$U_t$}: Time-conditional U-Net.
\FOR{each $l \in L$}
    \STATE Apply K-Means to partition $l$ into $C$ clusters.
    \STATE Tokenize class text to obtain word-frequency set $(w,freq)$.
    \FOR{each cluster}
        \STATE Calculate cluster center $z^c$ as image prototype
        \STATE Extract text prototype $T^c$ via Algorithm \ref{alg:text_prototype}.
    \ENDFOR
    \STATE $y = \tau_\theta(T^c)$ \COMMENT{Descriptive text embedding}
    \FOR{each $z^c$}
        \STATE $z^c_t \sim q(z_t^c | z^c)$ \COMMENT{Diffusion process}
        \STATE $\tilde{z}^c = U_t(\text{Concat}(z^c_t, y))$ \COMMENT{Denoising process}
    \ENDFOR
\ENDFOR
\STATE $S = D(\tilde{Z}^c)$ \COMMENT{Generate image}
\STATE \textbf{Output:} $S$: Distilled images
\end{algorithmic}
\caption{Dataset Distillation via Vision-Language Category Prototype}
\label{alg:1}
\end{algorithm}

\subsection{Outlier Removal}\label{B}
We apply Local Outlier Factor (LOF) \cite{alghushairy2020review}, a widely used unsupervised outlier detection method, to identify and remove data points with significantly lower density compared to surrounding samples. In comparison to other methods, LOF does not require ground truth, making it suitable for datasets with unknown distributions. We set two parameters for the LOF algorithm: $n\_neighbors=10$ is set for all datasets, while $contamination$ is adjusted based on the dataset characteristics. 
\subsection{Cross-Modal Information Distillation}\label{C}
\subsubsection{Image Prototypes}
Following the work of \cite{su2024d}, we first employ encoder $E$ to compress features in the latent space. Subsequently, we apply k-means clustering, a widely used unsupervised algorithm, to partition each category into a predefined number of clusters. The number of clusters is dynamically set according to the IPC. For instance, when IPC = 10, the number of clusters is set to 10. The cluster centers are then used as image prototypes, effectively capturing the representative visual features of each category. These prototypes provide compact yet informative features, facilitating more effective dataset distillation.

\subsubsection{Text Prototypes}
Effective dataset distillation should consider not only visual features but also semantic information, including details such as shape, posture, color, and other distinct features that may not be captured by the image features. Hence, we introduce the frequency-based prototype extraction method to obtain the text prototype for each cluster. This approach involves tokenizing the description text, filtering out non-representative words, and selecting the most representative sentence as the text prototype based on word frequency. The words $w$ with high frequency, appearing in more than $\beta$ proportion of the samples within the same category are excluded, as they are unlikely to characterize individual clusters. The procedure for calculating text prototypes is summarized in Algorithm \ref{alg:text_prototype}. 

First, all textual descriptions within a given class $l$ are tokenized to generate word-frequency set $(w,freq)$. Non-representative words $N$ are identified and removed:
\begin{equation}\label{eq1}
N = \left[ w \mid w, \text{freq} \in (w, \text{freq}) \text{ and } \frac{\text{freq}}{\text{len}(l)} > \beta \right].
\end{equation}
Next, the text data within each cluster are tokenized into individual words, and common stop words (e.g., ``is," ``the," ``of") are removed to generate a word-frequency set $(w_c,f_c)$. Nonrepresentative words are then excluded. The remaining words are ranked by frequency $f_c$ and the top-$k$ words are selected to generate representative words $Rw$:
\begin{equation}\label{eq2}
R_w = \{ w_c : f_c \mid w_c, f_c \in (w_c, f_c) \text{ and } w_c \notin N \}.
\end{equation}

Subsequently, the frequency of words in $Rw$ is then used as an importance score. Each text $t$ in the cluster is evaluated based on its matching score within the top-$k$ words, and the text with the highest score will be chosen as the text prototype.
\begin{equation}\label{eq3}
\text{Score}(t) = \textstyle\sum_{} Rw[w] \cdot \mathbb{I}(w \in t),
\end{equation}
\begin{equation}\label{eq4}
\mathbb{I}(w \in t) = \begin{cases} 
1, & \text{if } w \in Rw \\
0, & \text{otherwise}
\end{cases}.
\end{equation}
\subsection{Image Synthesis via LDM}\label{D}
Finally, we integrate the image and text prototypes into the latent diffusion model (LDM) to synthesize diverse and representative images. LDM employs a text encoder $\tau_\theta$ to project the descriptive text into the latent space, which then conditions the U-Net architecture to guide image synthesis, facilitating the fusion of cross-modal representation. For each cluster, the synthesis process is formulated as follows:
\begin{equation}
\text{output} = D\left(U_t\left(\text{Concat}\left(z_t^c, \tau_\theta(T^c)\right)\right)\right),
\end{equation}
where $D$ denotes the decoder and $z_t^c$ represents the cluster image prototype with noise. $T^c$ refers to the corresponding descriptive text prototype.

\begin{algorithm}[t]
\begin{algorithmic}[1]
\STATE \textbf{Input:} word-frequency set $(w,freq)$ of class $l$, descriptive texts $T$ of cluster
\STATE Identify nonrepresentative words $N$ based on Eq. \ref{eq1}: \\
\STATE Tokenize texts $T$ to obtain word-frequency set
$(w_c,f_c)$.

\STATE Select top$k$ representative words $R_w$ based on Eq. \ref{eq2}: \\

\STATE \textbf{Token}: Tokenize sentences into words.
\FOR{each text $t \in T$}
\STATE words = Token($t$)
\STATE calculate score via Eq. \ref{eq3};
\ENDFOR
\STATE \textbf{Output:}Select the top-score $t$ as the text prototype $T^c$
\end{algorithmic}
\caption{Generate text prototype for each cluster}
\label{alg:text_prototype}
\end{algorithm}

\begin{table*}[htbp]
\centering
\setlength{\arrayrulewidth}{0.4pt}

\resizebox{\textwidth}{!}{%
\begin{tabular}{ccccccccccccc}
\toprule
\textbf{IPC (Ratio)} & \textbf{Test Model} & \textbf{Random} & \textbf{K-Center} & \textbf{Herding} & \textbf{DiT} & \textbf{DM} & \textbf{IDC-1} & \textbf{GLaD} & \textbf{Minimax} & \textbf{D\textsuperscript{4}M} & \textbf{Ours} & \textbf{Full} \\ \midrule
 & ConvNet-6 & 24.3\scriptsize{±1.1} & 19.4\scriptsize{±0.9} & 26.7\scriptsize{±0.5} & \underline{34.2\scriptsize{±1.1}} & 26.9\scriptsize{±1.2} & 33.3\scriptsize{±1.1} & 33.8\scriptsize{±0.9} & 33.3\scriptsize{±1.7} & 29.4\scriptsize{±0.9} & \textbf{34.8\scriptsize{±2.4}} & 86.4\scriptsize{±0.2} \\
 10 (0.8\%) & ResNetAP-10 & 29.4\scriptsize{±0.8} & 22.1\scriptsize{±0.1} & 32.0\scriptsize{±0.3} & 34.7\scriptsize{±0.5} & 30.3\scriptsize{±1.2} & \underline{39.1\scriptsize{±0.5}} & 32.9\scriptsize{±0.9} & 36.2\scriptsize{±3.2} & 33.2\scriptsize{±2.1} & \textbf{39.5\scriptsize{±1.5}} & 87.5\scriptsize{±0.5} \\
   & ResNet-18 & 27.7\scriptsize{±0.9} & 21.1\scriptsize{±0.4} & 30.2\scriptsize{±1.2} & 34.7\scriptsize{±0.4} & 33.4\scriptsize{±0.7} & \underline{37.3\scriptsize{±0.2}} & 31.7\scriptsize{±0.8} & 35.7\scriptsize{±1.6} & 32.3\scriptsize{±1.2} & \textbf{39.9\scriptsize{±2.6}} & 89.3\scriptsize{±1.2} \\ \midrule
 & ConvNet-6 & 29.1\scriptsize{±0.7} & 21.5\scriptsize{±0.8} & 29.5\scriptsize{±0.3} & 36.1\scriptsize{±0.8} & 29.9\scriptsize{±1.0} & 35.5\scriptsize{±0.8} & - & \underline{37.3\scriptsize{±0.1}} & 34.0\scriptsize{±2.3} & \textbf{37.9\scriptsize{±1.9}} & 86.4\scriptsize{±0.2} \\
  20 (1.6\%) & ResNetAP-10 & 32.7\scriptsize{±0.4} & 25.1\scriptsize{±0.7} & 34.9\scriptsize{±0.1} & 41.1\scriptsize{±0.8} & 35.2\scriptsize{±0.6} & \underline{43.4\scriptsize{±0.3}} & - & 43.3\scriptsize{±2.7} & 40.1\scriptsize{±1.6} & \textbf{44.5\scriptsize{±2.2}} & 87.5\scriptsize{±0.5} \\
   & ResNet-18 & 29.7\scriptsize{±0.5} & 23.6\scriptsize{±0.3} & 32.2\scriptsize{±0.6} & 40.5\scriptsize{±0.5} & 29.8\scriptsize{±1.7} & 38.6\scriptsize{±0.2} & - & \underline{41.8\scriptsize{±1.9}} & 38.4\scriptsize{±1.1} & \textbf{44.5\scriptsize{±2.0}} & 89.3\scriptsize{±1.2} \\ \midrule
 & ConvNet-6 & 41.3\scriptsize{±0.6} & 36.5\scriptsize{±1.0} & 40.3\scriptsize{±0.7} & 46.5\scriptsize{±0.8} & 44.4\scriptsize{±1.0} & 43.9\scriptsize{±1.2} & - & \underline{50.9\scriptsize{±0.8}} & 47.4\scriptsize{±0.9} & \textbf{54.5\scriptsize{±0.6}} & 86.4\scriptsize{±0.2} \\
  50 (3.8\%) & ResNetAP-10 & 47.2\scriptsize{±1.3} & 40.6\scriptsize{±0.4} & 49.1\scriptsize{±0.7} & 49.3\scriptsize{±0.2} & 47.1\scriptsize{±1.1} & 48.3\scriptsize{±1.0} & - & \underline{53.9\scriptsize{±0.7}} & 51.7\scriptsize{±3.2} & \textbf{57.3\scriptsize{±0.5}} & 87.5\scriptsize{±0.5} \\
   & ResNet-18 & 47.9\scriptsize{±1.8} & 39.6\scriptsize{±1.0} & 48.3\scriptsize{±1.2} & 50.1\scriptsize{±0.5} & 46.2\scriptsize{±0.6} & 48.3\scriptsize{±0.8} & - & \underline{53.7\scriptsize{±0.6}} & \underline{53.7\scriptsize{±2.2}} & \textbf{58.9\scriptsize{±1.5}} & 89.3\scriptsize{±1.2} \\ \midrule
 & ConvNet-6 & 46.3\scriptsize{±0.6} & 38.6\scriptsize{±0.7} & 46.2\scriptsize{±0.6} & 50.1\scriptsize{±1.2} & 47.5\scriptsize{±0.8} & 48.9\scriptsize{±0.7} & - & \underline{51.3\scriptsize{±0.6}} & 50.5\scriptsize{±0.4} & \textbf{55.8\scriptsize{±1.7}} & 86.4\scriptsize{±0.2} \\
  70 (5.4\%) & ResNetAP-10 & 50.8\scriptsize{±0.6} & 45.9\scriptsize{±1.5} & 53.4\scriptsize{±1.4} & 54.3\scriptsize{±0.9} & 51.7\scriptsize{±0.8} & 52.8\scriptsize{±1.8} & - & \underline{57.0\scriptsize{±0.2}} & 54.7\scriptsize{±1.6} & \textbf{60.6\scriptsize{±0.3}} & 87.5\scriptsize{±0.5} \\
   & ResNet-18 & 52.1\scriptsize{±1.0} & 44.6\scriptsize{±1.1} & 49.7\scriptsize{±0.8} & 51.5\scriptsize{±1.0} & 51.9\scriptsize{±0.8} & 51.1\scriptsize{±1.7} & - & \underline{56.5\scriptsize{±0.8}} & 56.3\scriptsize{±1.8} & \textbf{60.3\scriptsize{±0.3}} & 89.3\scriptsize{±1.2} \\ \midrule
 & ConvNet-6 & 52.2\scriptsize{±0.4} & 45.1\scriptsize{±0.5} & 54.4\scriptsize{±1.1} & 53.4\scriptsize{±0.3} & 55.0\scriptsize{±1.3} & 53.2\scriptsize{±0.9} & - & 57.8\scriptsize{±0.9} & \underline{57.9\scriptsize{±1.5}} & \textbf{62.7\scriptsize{±1.4}} & 86.4\scriptsize{±0.2} \\
 100 (7.7\%) & ResNetAP-10 & 59.4\scriptsize{±1.0} & 54.8\scriptsize{±0.2} & 61.7\scriptsize{±0.9} & 58.3\scriptsize{±0.8} & 56.4\scriptsize{±0.8} & 56.1\scriptsize{±0.9} & - & \underline{62.7\scriptsize{±1.4}} & 59.5\scriptsize{±1.8} & \textbf{65.7\scriptsize{±0.5}} & 87.5\scriptsize{±0.5} \\
   & ResNet-18 & 61.5\scriptsize{±1.3} & 50.4\scriptsize{±0.4} & 59.3\scriptsize{±0.7} & 58.9\scriptsize{±1.3} & 60.2\scriptsize{±1.0} & 58.3\scriptsize{±1.2} & - & 62.7\scriptsize{±0.4} & \underline{63.8\scriptsize{±1.3}} & \textbf{68.3\scriptsize{±0.4}} & 89.3\scriptsize{±1.2} \\
\bottomrule

\end{tabular}%

}
\caption{Comparison of state-of-the-art methods on ImageWoof under various IPC settings and model architectures. All the results are obtained at a resolution of 256 $\times$ 256. The best results are marked as bold, and the second-best are underlined.}
\label{tab1}
\end{table*}

\section{Experiments}
\subsection{Datasets}
We evaluate the performance of our proposed method on both low-resolution and high-resolution datasets. For low-resolution datasets (32 $\times$ 32), we use CIFAR-10 \cite{krizhevsky2009learning} and CIFAR-100 \cite{krizhevsky2009learning}. For high-resolution data, we conduct experiments on ImageNet-1K \cite{deng2009imagenet} dataset and its subsets: ImageWoof, ImageNette \cite{howardsmaller}, ImageIDC \cite{kim2022dataset}, and ImageNet-100  \cite{tian2020contrastive}. ImageWoof is a challenging subset consisting of 10 fine-grained dog breed classes with high inter-class similarity. In contrast, ImageNette and ImageIDC contain 10 classes with lower inter-class similarity, making them easier to distinguish. Additionally, ImageNet-100 consists of 100 classes, and ImageIDC is derived from the first 10 classes of this dataset. 
\subsection{Implementation Details}
We conduct three independent trials with different seeds and report the average accuracy. We fine-tune the stable diffusion V1-5 model for each dataset using the generated image-text pairs. The batch size for fine-tuning is set to 8, and the training lasts 8 epochs. The resolution of the generated samples is set to 256 $\times$ 256 for ImageNet-1K subsets and 224 $\times$ 224 for the full ImageNet-1K dataset. For CIFAR-10 and CIFAR-100, the resolution is 32 $\times$ 32. For fair evaluation, we utilize the publicly available source code from \cite{gu2024efficient} to assess the performance of our method and report the top-1 accuracy on the original testing set. More implementation details are provided in the supplementary material.

\subsection{Comparison with the SOTA Methods}
We evaluate our method against the state-of-the-art approaches, including generative methods such as Minimax \cite{gu2024efficient}, $\text{D}^4\text{M}$ \cite{su2024d}, GLaD \cite{cazenavette2023generalizing}, and DiT \cite{peebles2023scalable,gu2024efficient}. Additionally, we compare our approach with decoupled distillation methods, including $\text{SRe}^2\text{L}$ \cite{yin2023squeeze} and RDED \cite{sun2024diversity}, as well as other techniques such as DM \cite{zhao2021datasetcondensation}, IDC-1 \cite{kim2022dataset}, Herding \cite{welling2009herding}, and K-Center \cite{sener2017active}. The mean and standard deviation of the results are reported. We reproduce Minimax \cite{gu2024efficient} method using the publicly available GitHub repository and conduct experiments under identical conditions to ensure a fair comparison.

\textbf{ImageWoof} We evaluate our method under varying IPC settings using three architectures: ConvNet-6, ResNetAP-10, and ResNet-18, as shown in Table \ref{tab1}. Across all settings and models, our method consistently outperforms the second-best method, demonstrating the robustness and adaptability of our approach. Notably, even with a low IPC (IPC = 10), our proposed method achieves 39.9\% accuracy with the ResNet-18 model, surpassing the second-best method by 2.6\%. As the IPC increases, the method still maintains its superiority, reaching 68.3\% accuracy at IPC = 100 with ResNet-18, which yields an improvement of 4.5\% compared to the $\text{D}^4\text{M}$ method. Furthermore, our method consistently achieves the best performance across various models—ConvNet-6, ResNetAP-10, and ResNet-18—demonstrating its robustness and generalization across different network architectures.

\textbf{ImageNette and ImageIDC} We assess our method using the ResNetAP-10 architecture under IPC 10, 20, and 50, as shown in Table \ref{tab2}. The results show that our method consistently outperforms all other approaches across varying IPC settings on both datasets. On the ImageNette dataset, our method significantly surpasses other methods, achieving a 4.1\% improvement in average accuracy. 

\begin{table}[htbp]
\centering

\resizebox{\linewidth}{!}{%
\begin{tabular}{cccccccc}
\toprule
& \textbf{IPC} & \textbf{Random} & \textbf{DiT} & \textbf{DM} & \textbf{Minimax} & \textbf{D\textsuperscript{4}M}  & \textbf{Ours} \\ \midrule
 & 10 & 54.2\scriptsize{±1.6} & 59.1\scriptsize{±0.7} & 60.8\scriptsize{±0.6} & 57.7\scriptsize{±1.2} &\underline{60.9\scriptsize{±1.7}} & \textbf{64.8\scriptsize{±3.6}} \\
 \textbf{Nette} & 20 & 63.5\scriptsize{±0.5} & 64.8\scriptsize{±1.2} & \underline{66.5\scriptsize{±1.1}} & 64.7\scriptsize{±0.8} & 66.3\scriptsize{±1.3} & \textbf{71.4\scriptsize{±0.5}} \\
  & 50 & 76.1\scriptsize{±1.1} & 73.3\scriptsize{±0.9} & 76.2\scriptsize{±0.4} & 73.9\scriptsize{±0.3} & \underline{77.7\scriptsize{±1.1}} & \textbf{81.2\scriptsize{±0.8}} \\ \midrule
 & 10 & 48.1\scriptsize{±0.8} & \underline{54.1\scriptsize{±0.4}} & 52.8\scriptsize{±0.5} & 51.9\scriptsize{±1.4} &50.3\scriptsize{±1.0} & \textbf{57.0\scriptsize{±1.4}}  \\
 \textbf{IDC} & 20 & 52.5\scriptsize{±0.9} & 58.9\scriptsize{±0.2} & 58.5\scriptsize{±0.4} & \underline{59.1\scriptsize{±3.7}}  & 55.8\scriptsize{±0.2} & \textbf{63.3\scriptsize{±1.2}} \\
  & 50 & 68.1\scriptsize{±0.7} & 64.3\scriptsize{±0.6} & 69.1\scriptsize{±0.8} & \underline{69.4\scriptsize{±1.4}} & 69.1\scriptsize{±2.4} & \textbf{71.9\scriptsize{±0.4}} \\ \bottomrule
\end{tabular}%
}
\caption{Comparison of the state-of-the-art methods on ImageNette and ImageIDC under various IPC settings. All the results are obtained on ResNetAP-10. The best results are marked as bold, and the second-best are underlined.}
\label{tab2}
\end{table}
\begin{table}[htbp]
\centering
\small
\begin{tabularx}{0.5\textwidth}{>{\centering\arraybackslash}c >{\centering\arraybackslash}X >{\centering\arraybackslash}X >{\centering\arraybackslash}X >{\centering\arraybackslash}X >{\centering\arraybackslash}X}
        \toprule
        IPC & $\text{SRe}^2\text{L}$  & RDED  & DiT & Minimax & Ours \\
        \midrule
        10  & 21.3\scriptsize{±0.6} & 42.0\scriptsize{±0.1} & 39.6\scriptsize{±0.4} & 44.3\scriptsize{±0.5} & \textbf{46.7\scriptsize{±0.4}} \\
        50  & 46.8\scriptsize{±0.2} & 56.5\scriptsize{±0.1} & 52.9\scriptsize{±0.6} & 58.6\scriptsize{±0.3} & \textbf{60.5\scriptsize{±0.2}} \\
        \bottomrule
    \end{tabularx}
    \caption{Performance comparison on ImageNet-1K.}
    \label{tab4}
\end{table}

\begin{table}[htbp]
\centering
\small
\begin{tabularx}{0.5\textwidth}{>{\centering\arraybackslash}c >{\centering\arraybackslash}c >{\centering\arraybackslash}X >{\centering\arraybackslash}X >{\centering\arraybackslash}X}
        \toprule
        Dataset & IPC & $\text{SRe}^2\text{L}$  & RDED & Ours \\
        \midrule
        \multirow{2}{*}{CIFAR-10}  & 10  & 29.3\scriptsize{±0.5} & 37.1\scriptsize{±0.3} & \textbf{39.0\scriptsize{±0.7}} \\
                                    & 50  & 45.0\scriptsize{±0.7} & 62.1\scriptsize{±0.1} & \textbf{63.2\scriptsize{±0.3}} \\
        \midrule
        \multirow{2}{*}{CIFAR-100} & 10  & 27.0\scriptsize{±0.4} & 42.6\scriptsize{±0.2} & \textbf{50.6\scriptsize{±0.7}} \\
                                    & 50  & 50.2\scriptsize{±0.4} & 62.6\scriptsize{±0.1} & \textbf{66.1\scriptsize{±0.3}} \\
        \bottomrule
    \end{tabularx}
    \caption{Performance comparison on CIFAR-10 and CIFAR-100.}
    \label{tab3}
\end{table}

The lower performance on ImageIDC compared to ImageNette may be attributed to the presence of two similar fine-grained classes in IDC: Saluki, and Doberman. Despite this, our method achieves notable performance improvements, with a 3.2\% increase in average accuracy, outperforming the state-of-the-art methods. The effectiveness of our method lies in its ability to simultaneously integrate both image and semantic information, unlike previous methods that only considered image features. 

\textbf{ImageNet-1K} We conduct experiments under IPC values of 10 and 50. All synthetic images are resized to 224 $\times$ 224 to ensure consistency with RDED \cite{sun2024diversity}. Table \ref{tab4} presents a performance comparison of various methods, including $\text{SRe}^2\text{L}$ \cite{yin2024squeeze}, RDED \cite{sun2024diversity}, DiT \cite{peebles2023scalable,gu2024efficient}, Minimax \cite{gu2024efficient}. The results indicate that our method consistently outperforms the others, achieving superior performance on large-scale datasets.

\textbf{CIFAR-10 and CIFAR-100} We also evaluate our method on two low-resolution datasets, CIFAR-10 and CIFAR-100, both with a resolution of 32 $\times$ 32. As shown in Table \ref{tab3}, our method surpasses $\text{SRe}^2\text{L}$ and RDED across different IPC settings on both datasets. Notably, on CIFAR-100 at IPC = 10, our method achieves a significant 8.0\% improvement over RDED \cite{sun2024diversity}. Our approach is especially effective for complex datasets like CIFAR-100, showcasing its robustness across datasets with various resolutions and complexities.

\subsection{Ablation Study}
As shown in Table \ref{tab5}, we conduct experiments with four configurations to evaluate various semantic strategies. These are assessed on two datasets, ImageIDC and ImageNette, under different IPC settings (10, 20, and 50).

Among the configurations, DCS consistently achieves the best performance across both datasets, except for ImageIDC at IPC-50. For instance, on ImageNette with IPC-50, DCS achieves an accuracy of 81.2$\pm$0.8\%, significantly outperforming other methods. Semantically closest sample descriptions provide highly relevant and context-rich information, enhancing synthesis quality and improving representativeness. In contrast, L+FK performs the worst overall. On ImageIDC with IPC-10, it achieves only 50.3$\pm$0.5\%, as the feature keywords lack logical relationships and are often disorganized, resulting in the synthesis of poor-quality images. The baseline (L) shows better performance than L+FK in most cases, as its simplicity avoids the noise introduced by poorly contextualized keywords. However, it lacks the semantic depth necessary for further improvements. GGS demonstrates moderate performance by introducing richer semantic context, leading to improved results compared to L and L+FK. Notably, it reaches 72.1$\pm$0.4\% on ImageIDC with IPC-50, surpassing DCS in terms of average accuracy. 

These results highlight the critical role of semantic information in improving the quality of the synthesized image. DCS consistently outperforms other methods, demonstrating the importance of context-rich descriptions to achieve superior synthesis performance.

\begin{table}[t]
\centering
\resizebox{0.5\textwidth}{!}{%
\begin{tabular}{cccccccc}
\toprule
\multirow{2}{*}{\parbox{1cm}{\centering \textbf{Semantic\\ Methods}}} & \multicolumn{3}{c}{\textbf{ImageIDC}} & \multicolumn{3}{c}{\textbf{ImageNette}} \\ \cmidrule(lr){2-4} \cmidrule(lr){5-7}
                              & \textbf{IPC-10}   & \textbf{IPC-20}   & \textbf{IPC-50}   & \textbf{IPC-10}   & \textbf{IPC-20}   & \textbf{IPC-50}   \\ \midrule
L                            & 54.1\scriptsize{±0.5}      & 61.5\scriptsize{±1.1}      & 71.2\scriptsize{±1.2}      & 60.1\scriptsize{±1.6}      & 69.7\scriptsize{±1.6}      & 76.6\scriptsize{±0.5}      \\
L+FK                         & 50.3\scriptsize{±0.5}      & 58.7\scriptsize{±2.1}      & 68.8\scriptsize{±2.3}      & 55.7\scriptsize{±1.4}      & 65.0\scriptsize{±4.6}      & 76.2\scriptsize{±0.9}      \\
GGS                        & 54.8\scriptsize{±3.1}      & 62.0\scriptsize{±1.9}      & \textbf{72.1\scriptsize{±0.4}} & 62.6\scriptsize{±2.7}      & 69.9\scriptsize{±1.3}      & 78.0\scriptsize{±0.1}      \\
DCS                        & \textbf{57.0\scriptsize{±1.4}} & \textbf{63.3\scriptsize{±1.2}} & 71.9\scriptsize{±0.4}      & \textbf{64.8\scriptsize{±3.6}} & \textbf{71.4\scriptsize{±0.5}} & \textbf{81.2\scriptsize{±0.8}} \\
\bottomrule
\end{tabular}%
}
\caption{Performance comparison on ImageNette and ImageIDC under various semantic methods: Label (L), Label + Feature Keywords (L+FK), GPT-Generated Sentences (GGS), and Descriptions of Closest Samples (DCS).}
\label{tab5}
\end{table}

\begin{table*}[h]
    \centering
    \scriptsize
    \begin{tabular}{>{\centering\arraybackslash}p{1.2cm} >{\centering\arraybackslash}m{2cm} >{\centering\arraybackslash}p{0.8cm} >{\centering\arraybackslash}p{5.5cm} >{\centering\arraybackslash}p{6cm}}
        \toprule
        Cluster (N) & Image & N-R & Feature Keyword & Text Prototype \\
        \midrule
        1 (145) & \begin{minipage}{2cm} \includegraphics[width=2cm]{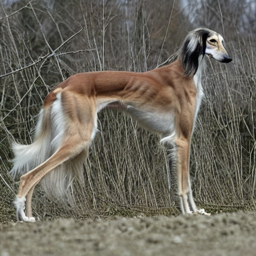}  \vspace{2pt} \end{minipage} & \begin{minipage}{1cm} \multirow{5}{*}{\shortstack{ \\  \\  \\  \\  \\  \\  \\  \\ \\  \\  \\   \\  \\ \\ \\  \\  white \\ gazelle \\ long \\ hound \\ dog \\ image \\ saluki \\ body \\ slender \\ color \\ tail \\ posture \\ distinctive}} \end{minipage}& \begin{minipage}{5.5cm} (large, 105), (shape, 94), (elegant, 80), (appearance, 77), (standing, 76), (curved, 75), (predominantly, 74), (markings, 74), (appears, 73), (head, 71), (brown, 67), (black, 62), (alert, 62), (legs, 60), (face, 59), (field, 59), (sleek, 56), (graceful, 53), (ears, 49), (grassy, 48), (possibly, 47), (breed, 42), (relaxed, 38), (pointed, 35), (features, 33), (adds, 33), (attentive, 31), (neck, 30), (coat, 29), (athletic, 27) \end{minipage} 
 &  \begin{minipage}{6cm} The Saluki, gazelle hound in the image is a large, slender dog with a long, lean body and a long tail. It has a distinctive shape, with a long head, pointed ears, and a long, curved muzzle. The dog is standing on a dirt road, and its posture appears alert and attentive. The color of the dog is predominantly white, with some brown markings on its face and body. The Saluki's features, such as its long legs and elegant posture, give it a graceful and athletic appearance. \vspace{2pt} \end{minipage}
 \\
        2 (105) & \begin{minipage}{2cm} \includegraphics[width=2cm]{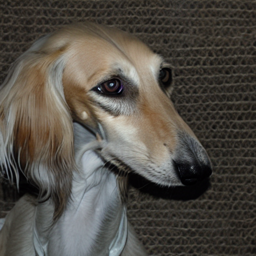} \vspace{2pt} \end{minipage} & & \begin{minipage}{5.5cm} (appearance, 78), (large, 76), (brown, 55), (shape, 53), (appears, 53), (curved, 49), (markings, 49), (predominantly, 48), (elegant, 46), (standing, 46), (ears, 45), (possibly, 43), (face, 43), (alert, 43), (legs, 42), (head, 41), (attentive, 41), (black, 33), (sleek, 32), (relaxed, 29), (breed, 29), (graceful, 27), (features, 26), (pointed, 25), (unique, 24), (suggests, 22), (eyes, 22), (looking, 22), (field, 21), (adds, 21) \end{minipage}
  &\begin{minipage}{6cm} The Saluki, gazelle hound in the image is small and slender, with a long and sleek body. It has a distinctive shape, with a long head, large ears, and a long tail. The dog is standing on a red carpet, and its posture appears to be relaxed and comfortable. The color of the dog is predominantly brown, with some black markings on its face and body. The Saluki's unique features, such as its long legs, long neck, and elegant appearance, make it an attractive and graceful breed.  \vspace{2pt} \end{minipage} \\
        3 (166) & \begin{minipage}{2cm} \includegraphics[width=2cm]{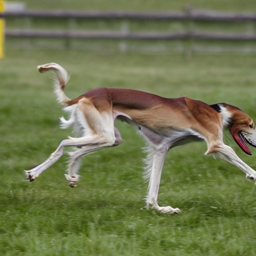} \vspace{2pt} \end{minipage} & & \begin{minipage}{5.5cm} (field, 130), (large, 112), (grassy, 112), (appearance, 107), (shape, 105), (brown, 96), (legs, 92), (predominantly, 82), (curved, 79), (markings, 79), (sleek, 79), (elegant, 78), (graceful, 77), (alert, 72), (running, 72), (head, 67), (ears, 64), (athletic, 62), (appears, 62), (standing, 59), (face, 56), (breed, 47), (black, 43), (focused, 41), (lean, 40), (possibly, 39), (adds, 38), (build, 36), (features, 34), (grass, 33) \end{minipage}
& \begin{minipage}{6cm}  The Saluki, gazelle hound in the image, has a slender and athletic build, with a long, lean body and a sleek coat. It has a distinctive shape, with a long, curved tail that extends downward. The dog's posture is energetic and graceful, as it is running swiftly across the grassy field. The Saluki's color is predominantly white, with some brown markings on its face and legs. The dog's eyes are open, and it appears focused on its surroundings, which adds to its overall dynamic appearance.  \vspace{2pt} \end{minipage}
 \\
        4 (120) & \begin{minipage}{2cm} \includegraphics[width=2cm]{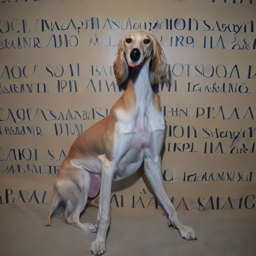} \vspace{2pt} \end{minipage} & & \begin{minipage}{5.5cm} (appearance, 92), (large, 90), (brown, 70), (elegant, 67), (head, 64), (shape, 62), (curved, 60), (appears, 54), (alert, 51), (legs, 50), (ears, 48), (standing, 48), (predominantly, 48), (markings, 45), (graceful, 41), (coat, 40), (relaxed, 38), (face, 37), (pointed, 37), (black, 31), (unique, 31), (looking, 30), (possibly, 30), (breed, 29), (features, 29), (eyes, 29), (adds, 29), (sleek, 29), (attentive, 28), (field, 27) \end{minipage}
 & \begin{minipage}{6cm} The Saluki, gazelle hound in the image is a large, white dog with a slender body and long legs. It has a distinctive shape, with a long head, a long neck, and a long tail. The dog appears to be well-groomed and well-behaved, standing on a blue carpet with a woman. The Saluki's posture is relaxed, and its color is predominantly white, with possibly some black markings on its face or body. The dog's overall appearance is elegant and graceful, which is typical of the Saluki breed. \end{minipage}
\\

        \bottomrule
    \end{tabular}
    \caption{An example of text prototypes corresponding to a ``Saluki, gazelle hound" class from dataset ImageIDC. Cluster (N) represents the cluster ID and sample size, while N-R denotes nonrepresentative words. Feature keywords are represented as (word, frequency) pairs.}
    \label{tab6}
\end{table*}

\subsubsection{Text prototype} 
The text prototype provides insights into the linguistic patterns associated with different clusters, highlighting both representative and nonrepresentative (N-R) characteristics, as shown in Table \ref{tab6}. In the ``Saluki, gazelle hound" class, nonrepresentative words appear in more than 70\% of the samples, including the class name itself and its common characteristics. For example, ``white", ``long" and ``slender" are classified as nonrepresentative words since they describe fundamental characteristics of the class: a slender dog with a long body and a coat that includes white. As these characteristics are prevalent across multiple clusters, they are unlikely to characterize individual clusters. 

\begin{figure*}[htbp] 
    \centering
    \includegraphics[width=0.85\textwidth]{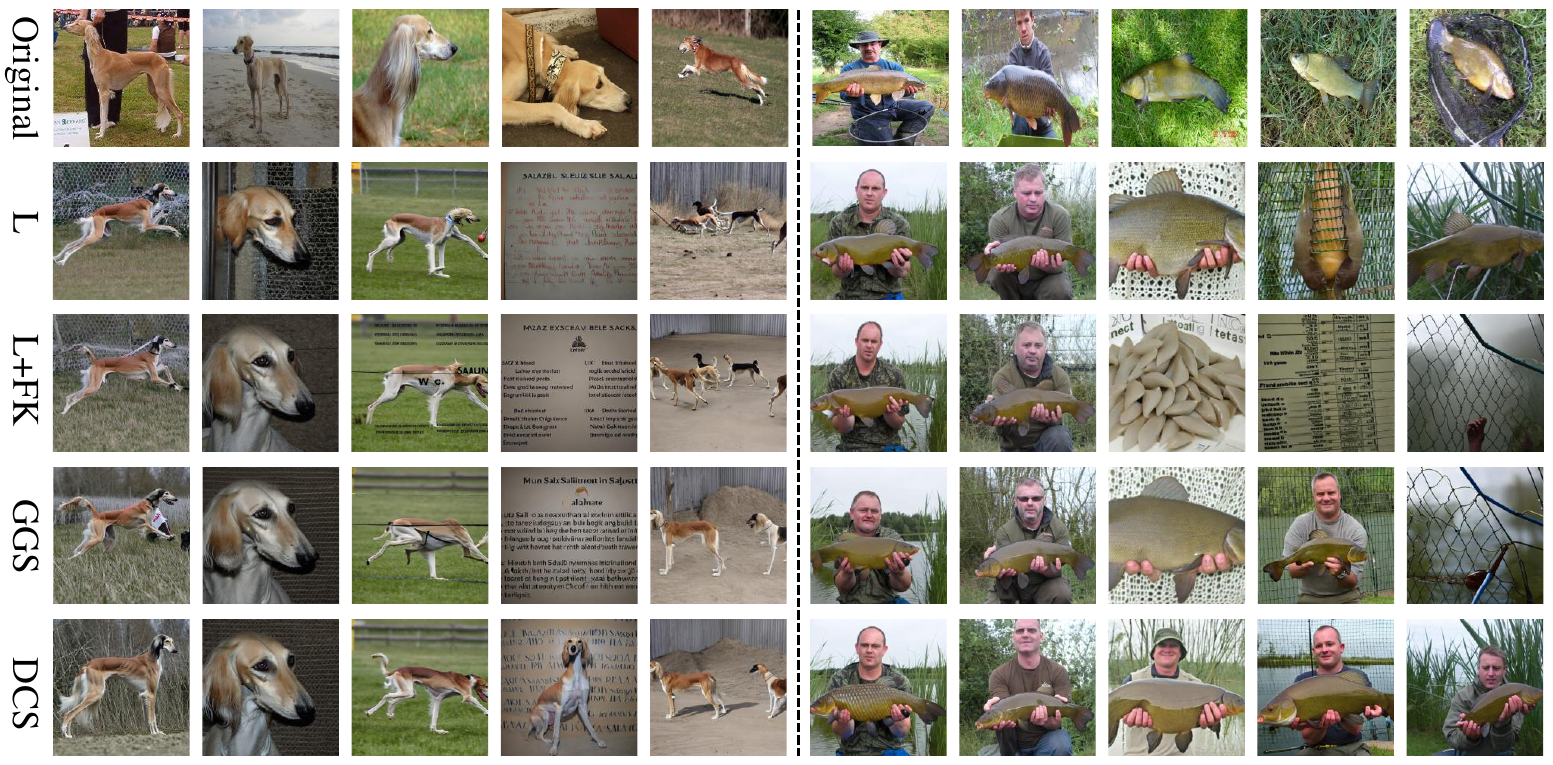} 
    \captionsetup{justification=raggedright} 
    \caption{Visualization of images generated using different semantic strategies. For each column, the images are generated using the same image prototype and random seed. In comparison, DCS produces images that are significantly more natural and logical.}
    \label{fig4}
\end{figure*}

Feature keywords are selected on the basis of their frequency, which serves as an importance score. In the feature keywords of Cluster 3 in Table \ref{tab6}, we observe that 112 out of 166 samples describe a ``grassy" background, 112 mention a ``large" target size, 96 feature the color ``brown" and 72 depict the action ``running". This indicates that nearly half of the brown dogs are running on the grass, which is consistent with the generated images. Moreover, compared to images generated using only labels shown in Fig. \ref{fig4}, our method produces more natural running postures and preserves detailed target features such as a curved tail. It also enhances logical consistency,  such as dogs with four legs rather than the five legs seen in the label-only images.

\begin{figure*}[]
\centering
\subfloat[] {\includegraphics[width=0.32\linewidth,scale=0.5]{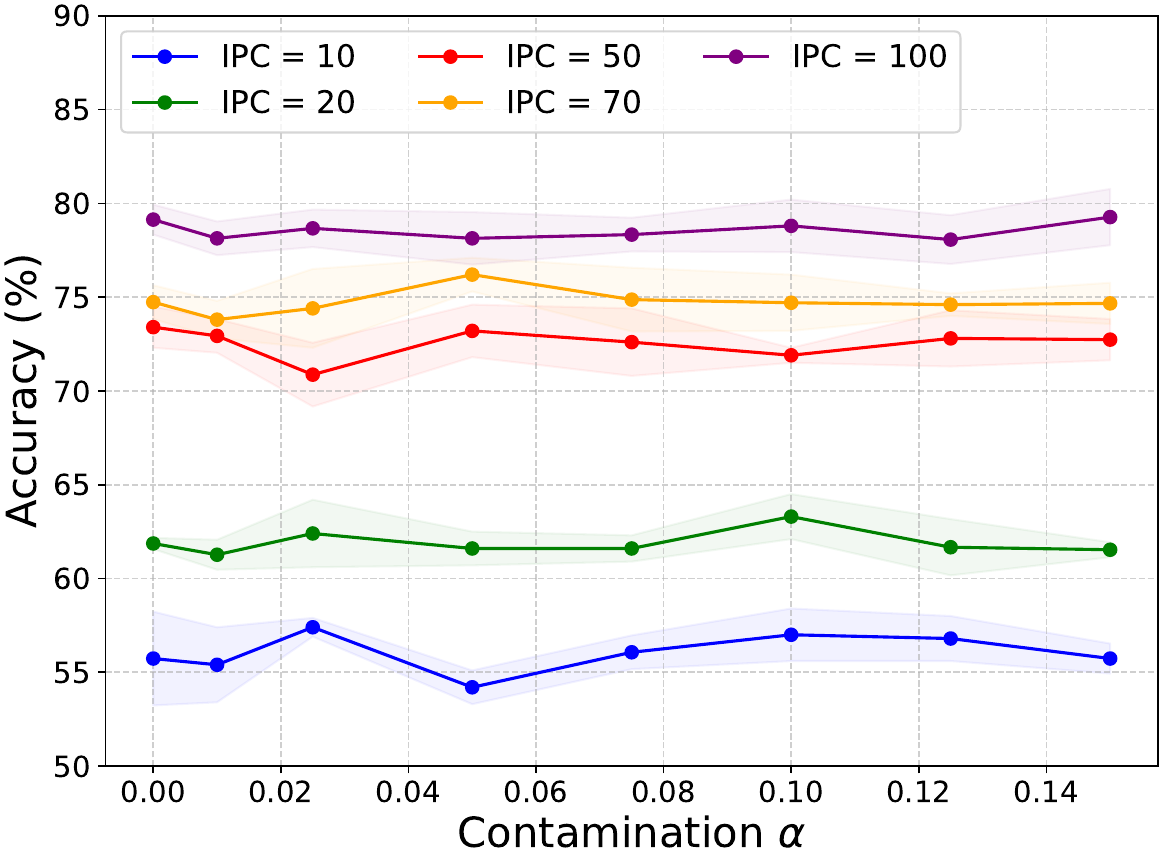}}\hspace{0.5em}%
\subfloat[]
{\includegraphics[width=0.32\linewidth,scale=0.5]{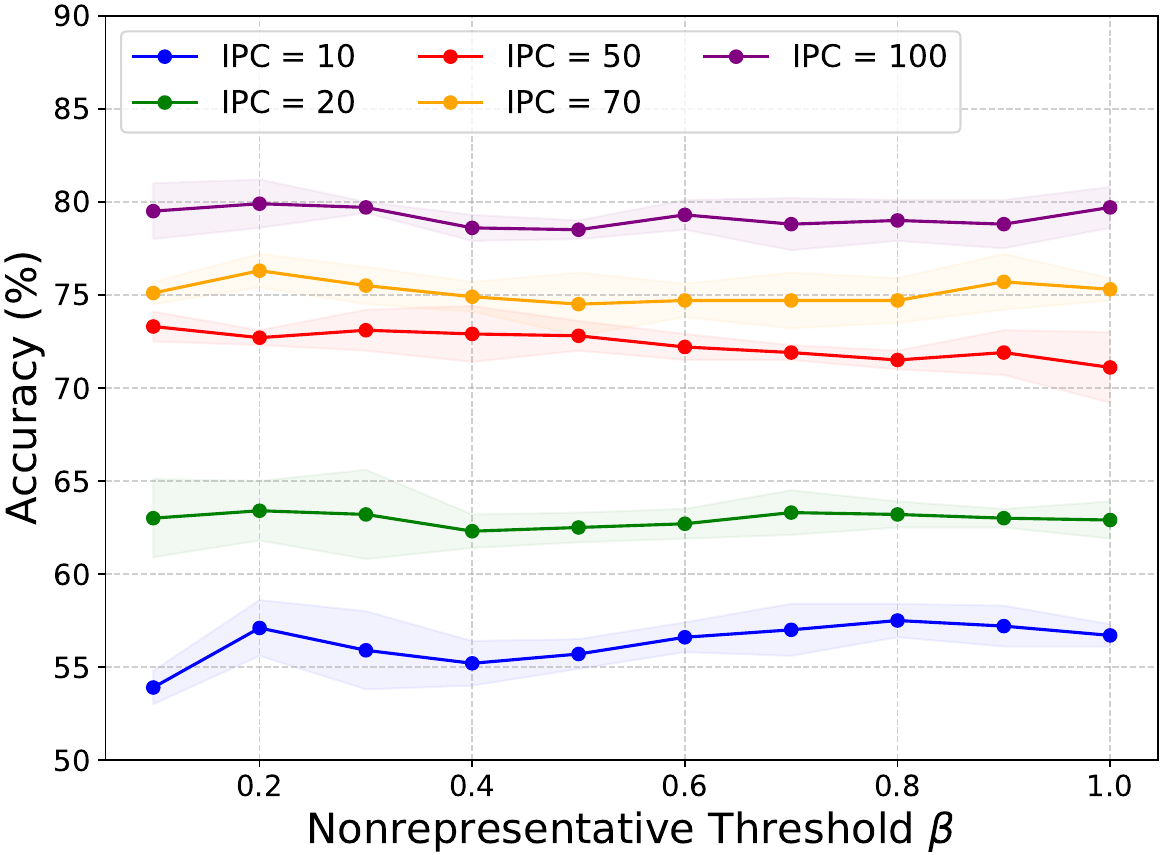}}\hspace{0.5em}%
\subfloat[]
{\includegraphics[width=0.32\linewidth,scale=0.5]{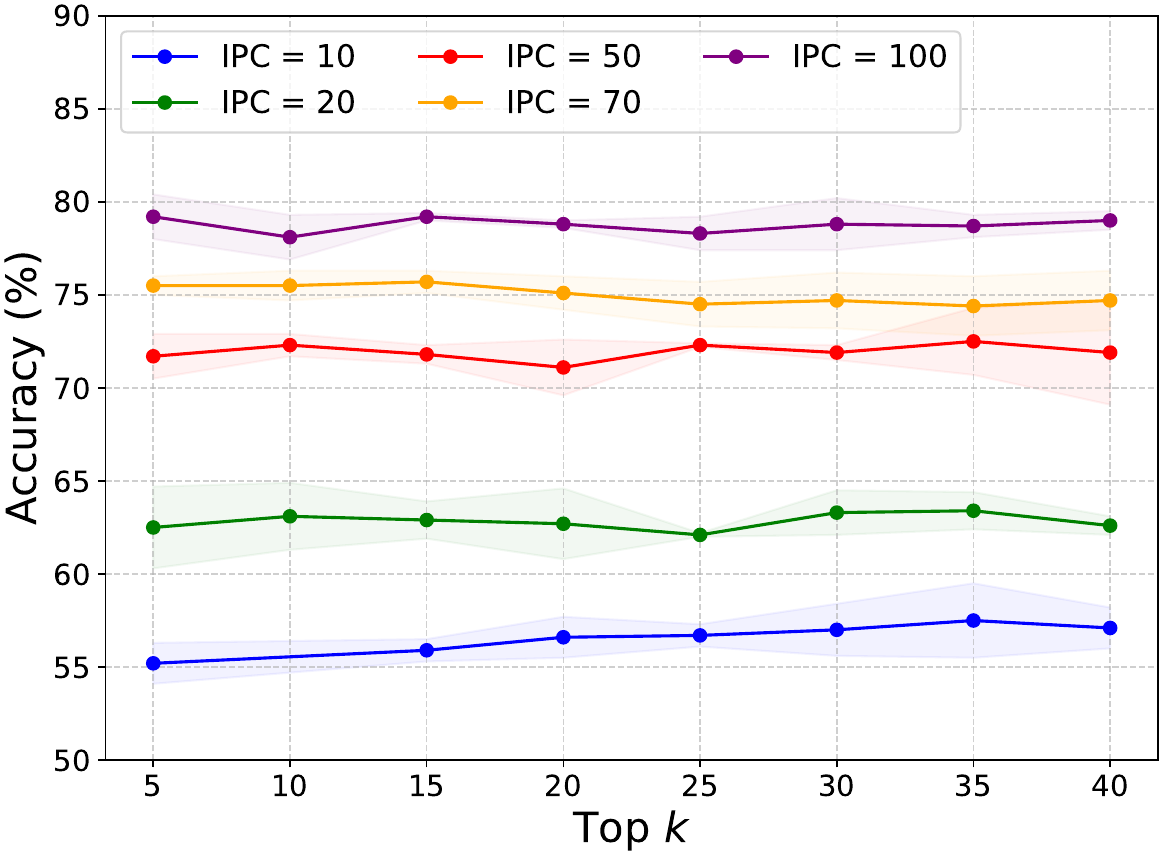}}\hspace{0.5em}%
\caption{Parameter Analysis of $\alpha$ (Contamination), $\beta$ (Nonrepresentative Threshold), and $k$ (Top-$k$ words) on ImageIDC.}
\label{fig:5}
\end{figure*}

\subsection{Visualization}
To evaluate the quality of the synthesized images, we compare samples generated using the same image prototype (corresponding to each column) across different semantic strategies, as illustrated in Fig. \ref{fig4}. The images on the left of the dashed line are sourced from ImageIDC and depict a Saluki, while those on the right are from ImageNette and represent a tench.
It can be observed that images generated by L, L+FK, and GGS all exhibit illogical outputs and the absence of objects. In contrast, DCS generates images that are more natural and structurally coherent, effectively preventing the absence of target objects.

In the Saluki case, L, L+FK, and GGS generate images with severe flaws, such as extra or missing limbs, while DCS consistently generates a logically coherent Saluki with the correct number of legs. Additionally, in the fourth column, only DCS successfully synthesizes images containing the target object, while the other methods fail to do so. Although L+FK contains more features than L, these words are unordered and unstructured, and they lack logical relationships, leading to misinterpretations. GGS generates sentences based on FK not encountered during the model's training, which may fail to provide the necessary context for accurate object generation. In contrast, DCS offers more detailed and context-rich descriptions, ensuring that the target object is consistently included in the generated image with all relevant features. More sample visualizations are provided in the supplementary material.

\subsection{Parameter Analysis}
We analyze the sensitivity of parameters $\alpha$ (Contamination), $\beta$ (Nonrepresentative Threshold), and $k$ (Top-$k$ words) on ImageIDC, as shown in Fig. \ref{fig:5} (a)-(c). The contamination parameter $\alpha$ has a significant impact on the performance. Models with lower IPC values (IPC = 10, 20, and 50) exhibit greater sensitivity to noisy data, resulting in more pronounced fluctuations in accuracy. In contrast, models with higher IPC values (IPC = 100) demonstrate stronger robustness, maintaining relatively stable accuracy. For $\beta$, except for IPC = 50, which exhibits a decreasing trend, other settings reach a peak at 0.2 before declining and stabilizing. Words appearing in more than 20\% of the samples in each class are classified as nonrepresentative words. As this threshold increases, high-frequency words within a class may be incorrectly selected as feature keywords for the cluster, reducing diversity. Regarding parameter $k$, as the value of top-$k$ increases, models with lower IPC values show a gradual increase in accuracy, reaching a maximum of 35, after which performance declines. This decline likely results from the inclusion of an increasing number of nonrepresentative words at higher top-$k$ values, leading to the selection of suboptimal text prototypes.

\section{Conclusion and Future Work}
In this work, we have proposed a novel dataset distillation method based on vision-language category prototypes. For the first time, we introduce text prototypes to complement image prototypes in dataset distillation, significantly enhancing the performance of the generated surrogate dataset. Compared to previous approaches, our method not only generates more logically coherent images containing target objects but also achieves outstanding performance across multiple benchmarks. By integrating the complementary strengths of visual and textual information, our approach provides a fresh perspective on dataset distillation, advancing the development of more efficient distillation techniques.

\textbf{Limitations and Future Works.} Our current work primarily focuses on classification tasks. In future research, we plan to extend our method to more complex vision tasks, such as object detection and segmentation, to evaluate its broader applicability. Additionally, we aim to explore alternative strategies for integrating text and image prototypes to further enhance the effectiveness of dataset distillation.

{
    \small
    \bibliographystyle{ieeenat_fullname}
    \bibliography{main}
}

\end{document}